\documentclass[11pt,letterpaper]{article}
\usepackage{cogsys}
\usepackage[T1]{fontenc}
\usepackage{times}
\usepackage[pdftex]{graphicx} 
\usepackage{tabularx} 
\usepackage{natbib}
\usepackage[utf8]{inputenc}
\ShortHeadings{Follow My Lead: Logical Fallacy Classification with Knowledge-Augmented LLMs}
              {O.\ Wang, T.\ Bansal, R.\ Bai, E.\ M.\ Chui, and L.\ H.\ Gilpin}

\DeclareUnicodeCharacter{2009}{\,}
\begin{document} 

\title{Follow My Lead: Logical Fallacy Classification with Knowledge-Augmented LLMs}
 
\author{Olivia Peiyu Wang}{pwang95@ucsc.edu}
\address{Computer Science and Engineering, UC Santa Cruz, Santa Cruz, CA 95064 USA}
\author{Tashvi Bansal}{tashvi.bansal@gmail.com}
\address{Monta Vista High School, Cupertino, CA 95014 USA}
\author{Ryan Bai}{ryanbai08@gmail.com}
\address{Canyon Crest Academy, San Diego, CA 92130 USA}
\author{Emily M. Chui}{emily.m.chui@gmail.com}
\address{Durham Academy Upper School, Durham, NC 27705 USA}
\author{Leilani H.\ Gilpin}{lgilpin@ucsc.edu}
\address{Computer Science and Engineering, UC Santa Cruz, Santa Cruz, CA 95064 USA}
\vskip 0.2in
 
\begin{abstract}
Large Language Models (LLMs) suffer from critical reasoning gaps, including a tendency to hallucinate and poor accuracy in classifying logical fallacies. This limitation stems from their default \emph{System 1} processing, which is \emph{fast and intuitive}, whereas reliable reasoning requires the \emph{deliberate, effortful System 2} approach \citep{kahneman2011thinking, li2025system}. Since full \emph{System 2} training is often prohibitively expensive, we explore a low-cost, instruction-based intervention to bridge this gap. Our methodology introduces a novel stepwise instruction dataset that decomposes fallacy classification into a series of atomic procedural steps (simple binary questions). We further augment this with a final verification step where models consult a relational knowledge graph of related fallacies. This procedural, rule-based intervention yields a significant improvement in LLM logical fallacy classification. Crucially, the approach also provides enhanced transparency into the LLMs' decision-making, highlighting a practical pathway for Neuro-symbolic architectures to address LLM reasoning deficits.
\end{abstract}

\section{Introduction} 
 
The emergence of Large Language Models (LLMs) has brought both challenges and opportunities. While LLMs demonstrate remarkable capabilities, they are susceptible to producing various forms of flawed output, including hallucinations and reasoning containing logical fallacies \citep{hong2023faithful,hong2024closer,liu2023evaluating,pan2023logic}. Logical fallacies are a specific class of errors that can make an argument invalid or deceptive due to a flaw in the structure or reasoning of an argument. For example, \emph{"Accent Fallacy"} can fundamentally alter the meaning and implications of a statement through the strategic manipulation of word emphasis. Encouragingly, the reasoning and learning abilities of LLMs have also shown significant potential in detecting hallucinations. \citep{kumar2024training} However, compared to factual errors, logical fallacies are more subtle; therefore, a more detailed and careful design is required to detect these incorrect patterns of reasoning. For example, research by Hong et al. shows that passing the definition of logical fallacies does not significantly improve the ability to detect logical fallacies. \citep{hong2024closer} Similarly, research has shown that LLMs have difficulty detecting specific types of logical fallacies, especially informal fallacies. \citep{hong2024closer} The capacity for LLMs to detect and classify logical fallacies is essential, given that these models frequently generate such flaws in their own generated reasoning. With appropriate design, this capability would enable self-revision to enhance the logical coherence of LLM outputs. \citep{saunders2022self} Furthermore, as online information, including political discourse \citep{luettgau2025conversational} and widespread misinformation, is often full of logical fallacies.  This shows the need to incorporate a detection mechanism for ensuring the fidelity of information retrieved and presented by LLMs. Beyond internal improvement, classifying fallacies could provide users with explicit explanations detailing why certain information constitutes misinformation, thereby fostering critical evaluation and preventing unwarranted trust in unsound sources. To address these limitations, this study proposes a symbolic approach for logical fallacy classification. Our methodology involves developing rule-based classification mechanisms that consist of stepwise instructions and relational graphs corresponding to each logical fallacy category presented in the FALLACIES dataset. \citep{hong2024closer} Subsequently, we conduct comprehensive quantitative and qualitative evaluations to assess LLM performance when guided by these structured instructions. Our research makes three key contributions to improving logical fallacy classification in Large Language Models:

\begin{enumerate}
\item \textbf{Development of the Atomic-Instruction-Dataset-for-Logical-Fallacies (AID-LF):} We systematically transform the existing FALLACIES dataset by decomposing complex logical fallacy descriptions into atomic, binary decision steps. Each fallacy is represented as a series of discrete yes/no questions that can be evaluated independently, accompanied by corresponding ground truth labels. This structured approach reduces classification complexity while maintaining comprehensive coverage of logical fallacy characteristics. The complete dataset is organized in JSON format for seamless integration with existing LLM pipelines\footnote{Our data is publicly available at \url{https://github.com/olivianxai/AID-LF}}. The notion of atomic instruction derives from the principles of decision procedures. These are algorithms characterized by their guaranteed termination with a binary outcome, either satisfiable or unsatisfiable, for any given formula.\citep{kroening2016decision} 

\item \textbf{Comprehensive Multi-Model Evaluation:} We conduct extensive empirical evaluation across multiple state-of-the-art language models, including Claude-Sonnet-4, ChatGPT-4o, ChatGPT-o4-mini, and Gemini-2.5-Flash. Our systematic comparison documents performance improvements against established baselines, with performance increases of 20.7\% for Claude-Sonnet-4, 3.4\% for both ChatGPT-4o and ChatGPT-o4-mini, and 8.7\% for Gemini-2.5-Flash.

\item \textbf{Prolog-Based Relational Graph Integration:} We develop and integrate Prolog-based relational graphs that model the structural connections between related logical fallacies.  This graph-based approach requires models to systematically examine interconnected fallacies before making final classifications, leading to more informed and accurate decisions. Our results demonstrate that structured rule-based guidance enables LLMs to achieve substantial performance improvements in the challenging domain of logical fallacy identification.
\end{enumerate}

\section{Related Work}

\textbf{Large Language Models in Logical Fallacy Detection and Classification: }
Recent research has explored various approaches to enhancing logical reasoning capabilities in Large Language Models, with particular attention to fallacy detection and self-correction mechanisms.
Kumar et al. demonstrated that structuring LLM reasoning processes to optimize self-evaluation and correction can yield improved performance on logic-based tasks in mathematics and code generation. \citep{kumar2024training} While their findings suggest promising potential for self-reflective reasoning, their investigation did not extend to testing logical fallacy detection nor classification within logical reasoning contexts.
Building upon similar self-sustaining methodologies, Jeong et al. developed a structured approach to mitigate logical fallacies in LLM outputs by systematically incorporating counterarguments, explanations, and objectives into model prompts. \citep{jeong2025large} Although this technique demonstrates effectiveness in reducing fallacious reasoning, it requires extensive manual curation of counterarguments, explanations, and objectives for individual statements, thus limiting its scalability and generalizability across diverse reasoning contexts.
Jin et al. advanced the field by creating a benchmark dataset focusing on climate-related fallacies and demonstrating that structure-aware classifiers for logical fallacy detection outperform existing LLM approaches. \citep{jin2022logical} However, their research failed to consider the similarities and disparities between different logical fallacies, leaving this potentially valuable application unexplored.
Lei et al. proposed an alternative structured approach utilizing logical structure trees to represent argumentative frameworks. \citep{lei2024boosting} Their methodology showed that providing LLMs with explicit logical tree structures with logical connective words as non-terminal nodes, while textual arguments as terminal nodes, significantly improved fallacy detection and classification performance. However, their approach primarily relies on connective word analysis, which may be insufficient for detecting certain categories of logical fallacies that manifest through more subtle, structural, or semantic patterns.
Most comprehensively, Hong et al. conducted extensive evaluations of LLM detection and classification capabilities using their FALLACIES dataset, which encompasses 232 distinct fallacy classes organized under formal and informal categories. \citep{hong2024closer} Despite the breadth of their evaluation, the best-performing model achieved only approximately 34\% average accuracy, indicating substantial room for improvement in current approaches.
These existing works collectively highlight both the importance of structured approaches to logical reasoning in LLMs and the significant challenges that remain in achieving reliable fallacy detection and classification capabilities.

\textbf{Instruction Following in Large Language Models: }In addition, current work on LLM instruction-following capabilities has revealed both the strengths and limitations of these models.
Qin et al. introduced a novel metric, Decomposed Requirements Following Ratio (DRFR), to quantitatively measure LLM instruction-following performance on a benchmark of 500 diverse instructions. DRFR involves decomposing each instruction within the benchmark into atomic criteria. LLMs' response to each question is thus evaluated on how many of the atomic criteria it satisfies, and the overall performance of the LLM on the benchmark is the number of criteria it has satisfied across all questions divided by the total number of atomic criteria in the benchmark. This evaluation method is a good representation of LLM performance on questions that involve multiple tasks; however, for questions centered on a single task that the LLM must handle, such as choosing one logical fallacy that fits a given example, assigning a score representing LLM performance requires different methods. Furthermore, although InFoBench's decomposed criteria advance interpretability, many of these ‘atomic’ checks are too vague. For example, one of the atomic criteria mentioned in InfoBench is ‘Is the generated questionnaire designed for hotel guests?’ This prompt offers little instruction on how an LLM should identify guest-specific content, thus undermining the validity of the DRFR. \citep{qin2024infobench}

White et al. proposed an entirely new, contamination-free benchmark, LiveBench. \citep{white2024livebench} Evaluating different LLMs such as Claude-Sonnet-3.5, GPT-4o, and Deepseek Coder v2. The instruction following capabilities of LLMs were found to have the highest performance out of all the tasks tested, at around 70\% accuracy. The impressive instruction following capabilities of LLMs give us the foundation for providing stepwise instructions to assist LLMs in logical fallacy classification.

Jiang et al. introduced FollowBench, a multi‑level, fine‑grained constraints benchmark that evaluates LLMs by incrementally adding one atomic constraint at each level. \citep{jiang2023followbench} They found that incrementally adding constraints reduced instruction-following performance, with the most advanced models reliably following instructions contain only around three constraints. Their findings demonstrate that simpler, atomic instructions may yield more consistent results.

\textbf{Prompting Best Practices for Large Language Models: }Kim et al. tested various methods to prompt LLMs on the same task. Expert-role descriptions, longer and more detailed instructions, step-by-step instructions, and providing examples (multi-shot prompting) were all found to improve LLMs' performance. \citep{kim2023better} In support of the role-based approach, Kong et al. found that role-play prompting (e.g., "you are an ...") improved LLM performance in comparison to standard zero-shot prompting. \citep{kong-etal-2024-better}
Zamfirescu-Pereira et al. showed that writing prompts with demarcations, rephrasing the prompts in the format of code (such as Jinja), and repeating instructions can improve LLM outputs. \citep{zamfirescu2023johnny}
Prior work also cautions that LLMs often misinterpret negated instructions and can even suffer performance drops when prompts are phrased negatively. \citep{jang2023can,li2023maqa,lou2024large,mishra2021reframing}
In this study, we adopt all prompting best practices identified in the literature review to optimize output quality.

\textbf{Logical Reasoning and Prolog: } Prolog, standing for "PROgramming in LOGic," was developed as a language for the logic programming paradigm. Its declarative nature and powerful built-in inference engine quickly established it as a primary tool for symbolic AI research. Historically, Prolog has been instrumental in the development of Expert Systems and Knowledge Representation \citep{bratko1990prolog,clark1980prolog,hayes1983building} and in systems for Automated Reasoning. \citep{lund2022verified}

While other paradigms have since dominated the field, a renewed interest in logical programming has emerged with the rise of Neuro-symbolic AI. \citep{belle2025relevance, chen2025comparative, colelough2025neuro, garcez2023neurosymbolic, vakharia2024proslm, saccon2024prolog, tan2024thought} This modern approach seeks to combine the strengths of neural networks with the formal, verifiable reasoning of symbolic methods. This revival stems from the paradigm's inherent ability to provide rule-based constraints and symbolic grounding, which can be integrated with neural network models. This integration offers a path toward verifiable AI systems and is particularly effective in addressing the problem of hallucinations.

In this context, Prolog is uniquely suited for our work. Its declarative syntax is ideal for defining and querying the complex relationships between different logical fallacies. By leveraging these core features, we can create a robust and efficient query system for related logical fallacies.
\section{Methodology}

\subsection{Stepwise Instructions Dataset Construction}
To develop our structured stepwise instructions dataset, we extracted logical fallacies and their corresponding descriptions from the FALLACIES dataset. \citep{hong2024closer} We then transformed these descriptions into systematic, step-by-step instructions by decomposing them into atomic binary questions that can be answered with "yes" or "no" responses. Afterwards, the authors acted as human annotators and performed three rounds of annotations to verify and polish each instruction, add ground-truths, and operations references to provide LLMs with comparative benchmarks during the classification process. The procedural steps, ground truths, and operations references were consolidated into a structured JSON file designated as "final\_instructions.json", which we will refer to as the "stepwise instructions" throughout this paper. In Appendix B, we include an example of "Accent Fallacy" before and after the transformation.  
\begin{figure}[t]
\vskip 0.05in
\begin{center}
     \includegraphics[width=1\linewidth]{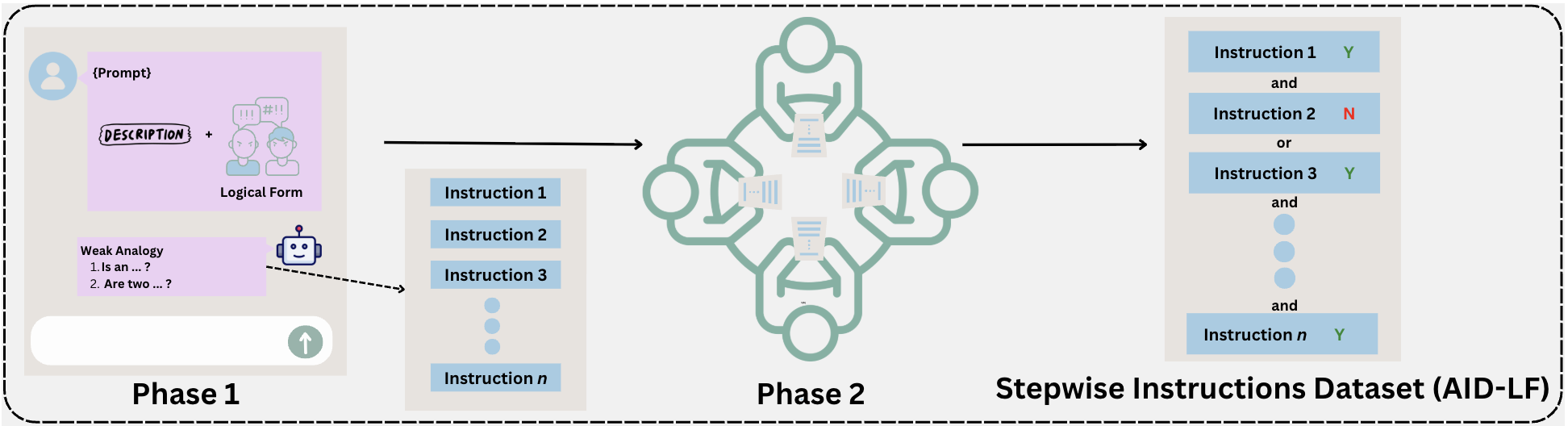}
     \caption{Visualization of Stepwise Instructions Dataset Pipeline. This figure illustrates the two-phase pipeline used to create the Atomic-Instruction-Dataset-for-Logical-Fallacies (AID-LF). Phase 1 involved using large language models (LLMs) to decompose the descriptions and logical forms of 232 fallacies into atomic, actionable steps. Phase 2 consisted of a rigorous human annotation process, where authors performed three rounds of verification and cross-checking. The annotations ensured that the instructions accurately reflected the original fallacy definitions, minimized redundancy, and provided complete coverage.}
\end{center}
\vskip -0.2in
\end{figure}
 
\subsection{Baseline Establishment}

We established our baseline by providing LLMs with the descriptions and logical forms of all logical fallacies, then testing their ability to identify the fallacy present in a given test statement. The example statements were curated by combining the first example listed for each of the 232 fallacies in the FALLACIES database, and these same statements were used across all approaches. Our evaluation employed three state-of-the-art language models: Claude-Sonnet-4, \citep{anthropic2025_claude4_systemcard} ChatGPT-4o, \citep{achiam2023gpt} and Gemini-2.5-Flash. \citep{team2023gemini}
\begin{figure}[t]
\vskip 0.05in
\begin{center}
    \includegraphics[width=0.5\linewidth]{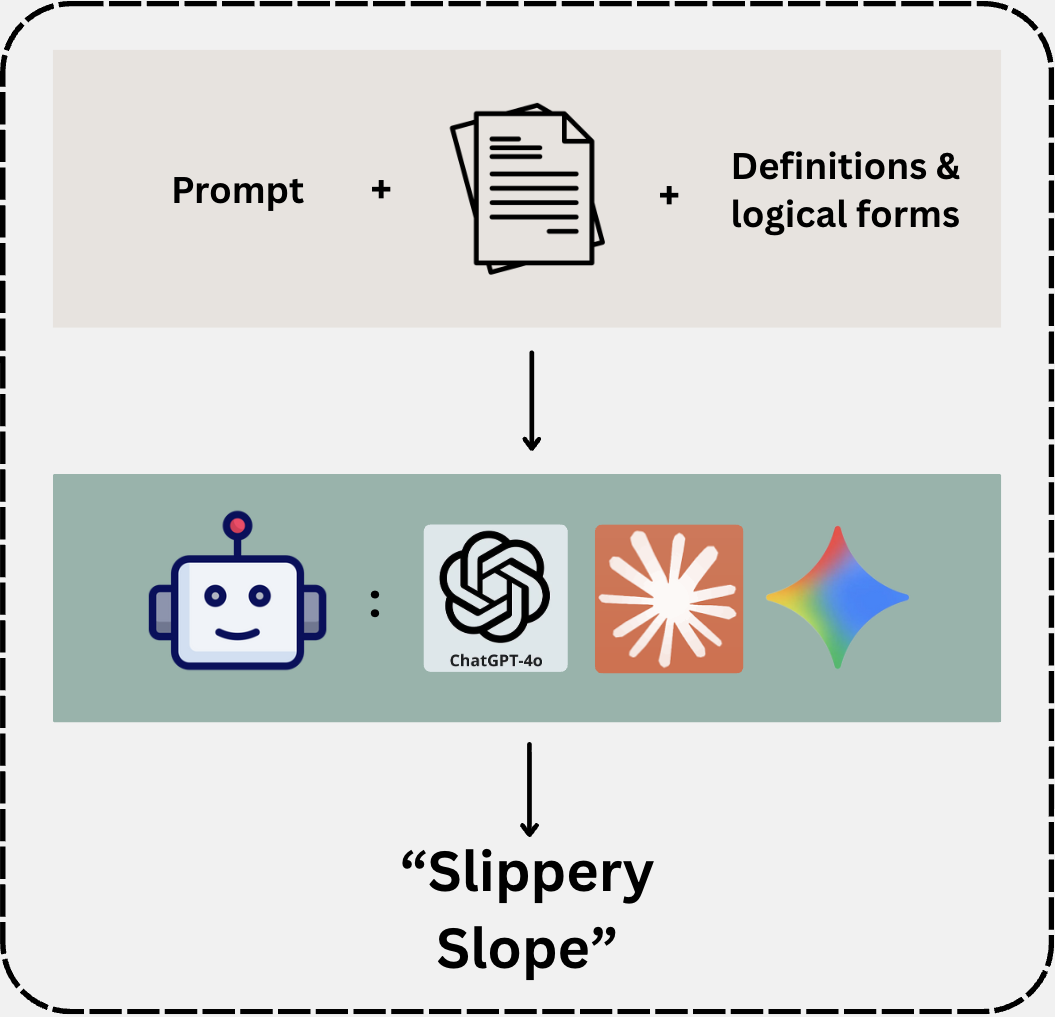}
     \caption{Baseline Establishment Process. For the baseline, large language models (LLMs) were provided with the descriptions and logical forms of all 232 fallacies, along with a statement for classification. The LLMs were instructed to select the most appropriate fallacy.}
\end{center}
\vskip -0.2in
\end{figure}

\subsection{Three-tiered Hierarchical Classification}
While Hong et al. established a hierarchy of logical fallacies, they did not explore hierarchical classification methodologies. \citep{hong2024closer} To investigate whether such a hierarchical organization could enhance logical fallacy classification performance, we implemented a three-tiered hierarchical classification approach.
At the first tier, LLMs determine whether a given statement contains a formal or informal logical fallacy, with explicit definitions of both categories provided as context.
At the second tier, classification proceeds based on the initial determination. Statements classified as formal fallacies are further categorized into four subcategories: "Proposition," "Quantification," "Syllogism," and "Probability." Those identified as informal fallacies are classified among five subcategories: "Ambiguity," "Inconsistency," "Irrelevance," "Insufficiency," and "Inappropriate Presumption." Complete definitions for all second-tier categories are provided to the models. Models are also permitted to revise their initial classification if deemed necessary.
At the third tier, LLMs perform the most granular classification based on their previous selections, again with comprehensive definitions supplied. Models retain the option to modify their prior classifications at this stage as well.
\begin{figure}[t]
\vskip 0.05in
\begin{center}
     \includegraphics[width=0.9\linewidth]{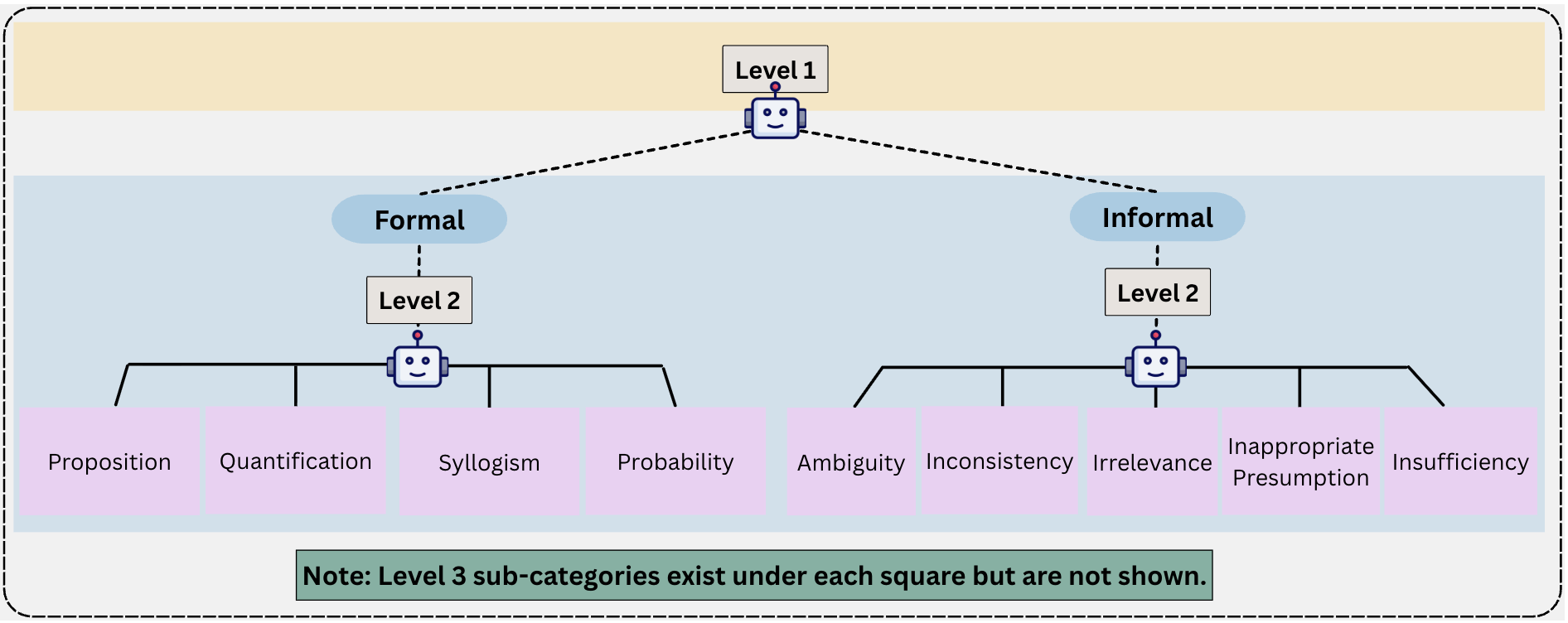}
     \caption{Three-tiered Hierarchical Classification of Logical Fallacy. The hierarchical classification process proceeded in three distinct levels.
    \textbf{Level 1:} The LLMs were initially prompted to classify a statement as either a Formal or Informal Fallacy, using only the definitions of these two categories.
    \textbf{Level 2:} Based on the first-level classification, the models were then supplied with definitions for the corresponding subcategories (e.g., "Proposition," "Quantification," "Syllogism" for formal fallacies, or "Ambiguity," "Inconsistency," "Irrelevance," "Insufficiency," "Inappropriate Presumption" for informal fallacies). The LLMs were given the option to revise their initial classification.
    \textbf{Level 3:} For the final, most granular classification, the models were prompted to select the specific fallacy from a detailed list that included each fallacy's description and logical form. At this stage, they could also revise any prior decisions.}
\end{center}
\vskip -0.2in
\end{figure}

\subsection{Stepwise Instructed Classification}
To evaluate the effectiveness of our stepwise instructional approach, we provided the stepwise instructions to the LLMs and requested classification of logical fallacies within test statements by following the prescribed steps and comparing results against the established ground truths. Following prompt‑engineering best practices, each prompt was initiated with an expert-role descriptor. \citep{kim2023better,kong-etal-2024-better} Subsequently, prompts were structured using explicit headings and labels that delineated the knowledge-base architecture, classification procedures, and critical requirements. \citep{white2023prompt} The critical requirements incorporated iterative instructions to emphasize the necessity of strict compliance with the specified output format. \citep{zamfirescu2023johnny} To enhance prompt effectiveness, example text and desired LLM output formats were enclosed within explicit delimiters. \citep{chen2025unleashing} Finally, to ensure consistency across all evaluations, this structured prompt framework was uniformly applied to all models tested. 
\begin{figure}[t]
\vskip 0.05in
\begin{center}
    \includegraphics[width=0.6\linewidth]{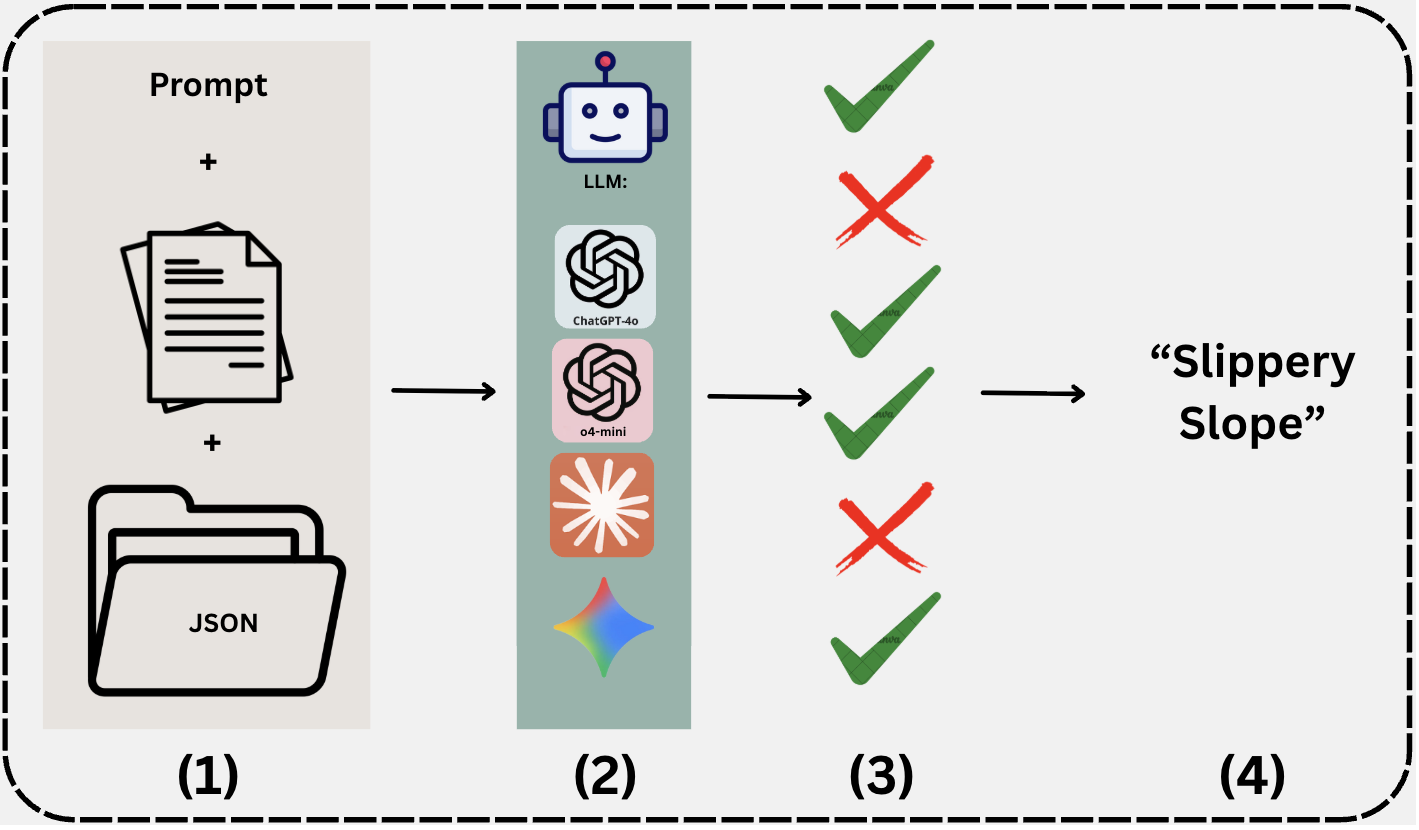}
     \caption{Stepwise Instruction Classification Process. The LLMs were supplied with the stepwise instruction dataset and the statement to be classified. They were instructed to execute the steps for each fallacy and return the first fallacy for which all steps matched the ground truth.}
\end{center}
\vskip -0.2in
\end{figure}

\subsection{Instruction-Guided Classification with Relational Graphs}
Our relational graphs were implemented with Prolog. Although Prolog was not explicitly designed for knowledge representation (KR), its capabilities are sufficient for establishing the necessary relationships between related logical fallacies, making it an appropriate choice for this study. The selection of Prolog is strategically motivated by its potential to form the basis for future work, enabling the integration of detailed, Prolog-based representations for each fallacy and facilitating the automatic generation of classification deductions and proofs.
Our baseline results revealed significant overlap among several logical fallacies. We observed that LLMs frequently misclassified "Contextomy" as the "Accent Fallacy", and vice versa. This confusion is notable because the two fallacies have distinct definitions: "Contextomy" manipulates meaning through altering the context, while "Accent Fallacy" does so by applying emphasis on specific words. The models' inability to recognize this key difference indicates a potential weakness that could be addressed by developing more precise classification frameworks. To capture these intricate inter-fallacy relationships, we developed relational graphs using Prolog, a symbolic programming language well-suited for logical reasoning tasks, based on the misclassification results from our baseline experiments. 

To evaluate the combined efficacy of stepwise instructions and symbolic relational graphs, we implemented a comprehensive three-phase classification protocol. We provided both the stepwise instructions as well as the relational graphs to the LLMs and requested classification of logical fallacies within the test statements. Before giving a final classification, LLMs were instructed by the prompt to complete a three-step process. 
Step 1 requires LLMs to read the stepwise instruction file, execute each step specified in the fallacy's instruction set, and compare the execution results to the "ground\_truths" and find initial matches.
Step 2 requires LLMs to read the relational graphs and, for every initial match identified in Step 1, find all related fallacies from the relational graphs. Then, LLMs are requested to return to the stepwise instructions and execute all the step-by-step analysis similarly to Step 1, and document all possible matches.
Step 3 requires LLMs to compare all results from Steps 1 and 2 and select the fallacy that best fits the statement. The decision is based on the strength of the match with ground truths, the quality of the step-by-step analysis, and the logical consistency across all steps.
The results of these approaches are presented and analyzed in the subsequent sections.

\begin{figure}[t]
\vskip 0.05in
\begin{center}
    \includegraphics[width=1\linewidth]{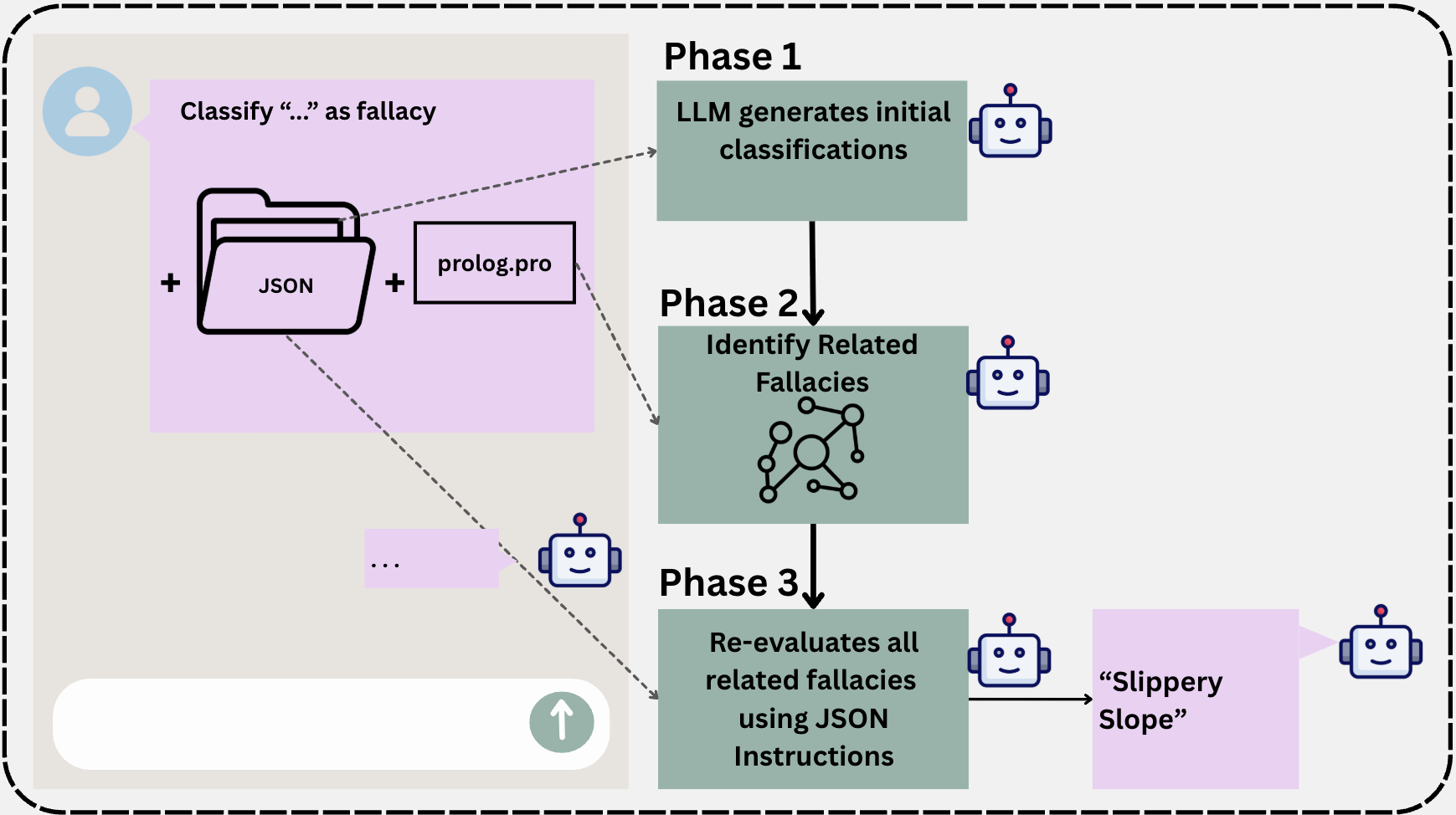}
     \caption{Instruction-Guided Classification with Relational Graphs}
\end{center}
\vskip -0.2in
\end{figure}

It should be noted that we also included ChatGPT-o4-mini in our experiments using the "Stepwise Instructed Classification" and "Symbolic Relational Graphs Assisted Classification" methods. This was done to rule out the possibility that the superior performance of Claude-Sonnet-4 was due to data contamination, given that ChatGPT-4o was released approximately one year before the other models. As the performance of ChatGPT-4o and ChatGPT-o4-mini was nearly identical, we opted to only include the results from the ChatGPT-4o experiments for clarity and brevity.
\section{Results} 

\subsection{Baseline Results}
Claude-Sonnet-4, Gemini-2.5-Flash, and ChatGPT-4o achieved comparable performance in the baseline evaluation, achieving 42.2\%, 43.5\%, and 44.0\% accuracy, respectively. These baseline results align with the findings reported by Hong et al. \citep{hong2024closer}

\subsection{Three-tiered Hierarchical Classification Results}
Given the hierarchical structure of this classification methodology, a classification was considered correct only if the LLMs identified the correct label across all three levels. The three models demonstrated comparable performance under the hierarchical classification approach, with Claude-Sonnet-4 and ChatGPT-4o both achieving 25.9\% accuracy, while Gemini-2.5-Flash achieved 24.6\% accuracy. Notably, all three LLMs exhibited diminished performance compared to the baseline when employing the hierarchical classification approach.
Upon analyzing the results, we identified that the majority of classification errors occurred at the second hierarchical level. Despite the structural provision allowing LLMs to revise their previous classifications, such revisions occurred infrequently. Instead, models typically maintained consistency with their prior classifications and proceeded to select options that aligned with their previous determinations. Consequently, when an incorrect classification was made at the second level, models rarely recovered to achieve overall classification accuracy.
\subsection{Stepwise Instruction Results}
The stepwise instruction results present the most compelling findings. Claude-Sonnet-4 demonstrated superior performance compared to the baseline, achieving 55.2\% accuracy, while Gemini-2.5-Flash exhibited diminished performance of only 39.2\% accuracy, and ChatGPT-4o achieved the lowest performance, which is 36.6\% accuracy, among the three models.

\subsection{Instruction-Guided Classification with Relational Graphs Results}
Across all models, the Instruction-Guided Classification with Relational Graphs approach yielded the highest performance. Claude-Sonnet-4 notably outperformed the other models with an accuracy of 62.9\%, while Gemini-2.5-Flash achieved 52.2\%. Although it showed the least performance improvement, ChatGPT-4o still reached an accuracy of 47.8\%.

\begin{table}[h!]
\begin{tabularx}{\textwidth}{|l|X|X|X|}
\hline
\textbf{Evaluation Type} & \textbf{Claude-Sonnet-4 Accuracy} & \textbf{ChatGPT-4o Accuracy} & \textbf{Gemini-2.5-Flash Accuracy} \\
\hline
Baseline Classification & 42.2\% & 44.0\% & 43.5\% \\
Hierarchical Classification & 25.9\% & 25.9\% & 24.6\% \\
Stepwise Classification & 55.2\% & 36.6\% & 39.2\% \\
Enhanced Classification & 62.9\% & 47.8\% & 52.2\% \\
\hline
\end{tabularx}
\caption{Comparison of Accuracy Across Models and Classification Approaches}
\label{tab:classification_results}
\end{table}

\begin{figure}[t]
\vskip 0.05in
\begin{center}
     \includegraphics[width=1\textwidth]{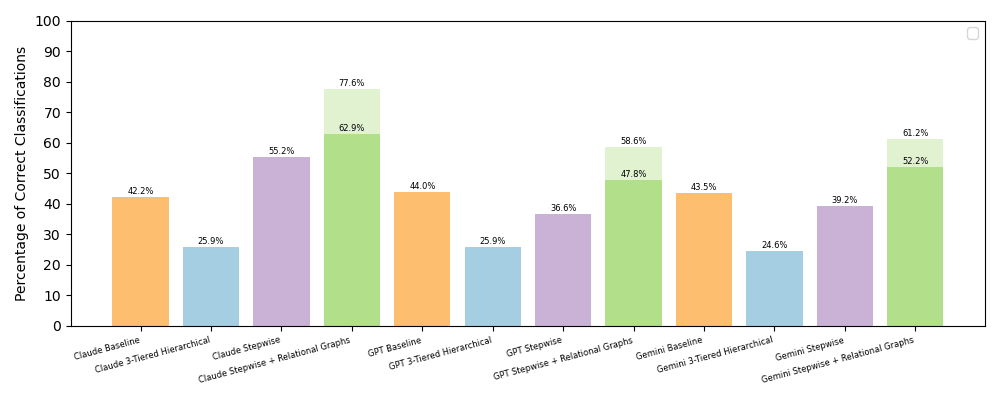}
     \caption{Performance Comparison Across Models and Different Approaches. The Stepwise + Relational Graphs method (dark green) achieved the highest accuracy. Light green shows the percentage of misclassified statements where the correct label was the second-best prediction. Other colors represent different approaches as labeled.}
\end{center}
\vskip -0.2in
\end{figure}
\section{Discussion}
\textbf{Methodological Shortcomings} 
In our Three-tiered Hierarchical Classification method, most errors occurred at the second level. LLMs rarely corrected these errors, instead maintaining consistency with their initial, incorrect classifications. This lack of self-correction meant that a mistake at the second level almost always led to an incorrect final classification.

\textbf{LLM Interpretation and Misclassification} Below are the most notable errors and their causes from the experiments:
\vskip 0.05in
\cbullet
Speaker Intention: Across all models (ChatGPT-4o, Claude-Sonnet-4, Gemini-2.5-Flash, and ChatGPT-o4-mini), we observed a tendency to misinterpret speaker roles in multi-person interactions, leading to incorrect classifications. For example, in a "fallacy of accomplishment" case, LLMs mistakenly identified the subject of the argument as the arguer, resulting in a wrong classification.
\cbullet 
Ad Hominem Fallacies: The inability to distinguish between neutral, satirical, or ironic statements and genuine attacks led to high misclassification rates for "Ad Hominem" fallacies, particularly for Claude-Sonnet-4.
\begin{figure}[t]
\vskip 0.05in
\begin{center}
     \includegraphics[width=1\textwidth]{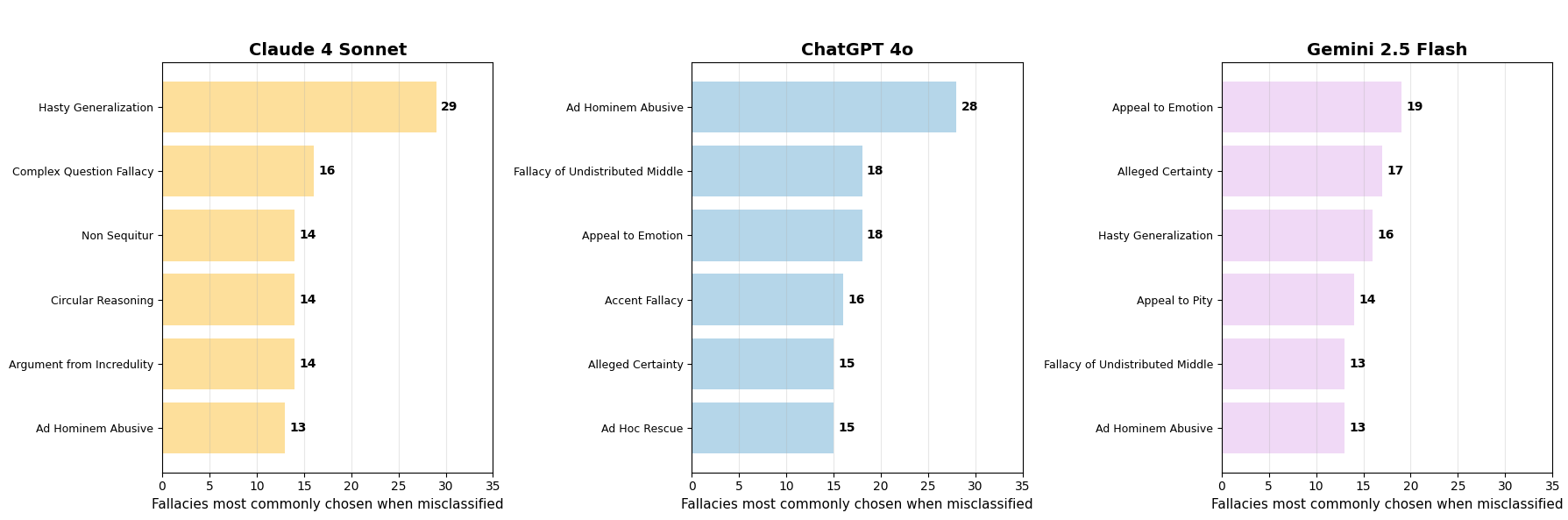}
     \caption{LLM Top Fallacy Misclassifications. Among the misclassified statements, Claude-Sonnet-4's top three misclassifications are "Hasty Generalization", "Complex Question Fallacy", and "Non Sequitur". ChatGPT-4o's top three misclassifications are "Ad Hominem Abusive", "Fallacy of Undistributed Middle", "Appeal to Emotion". Gemini-2.5-Flash's top three misclassifications are "Appeal to Emotion", "Alleged Certainty", and "Hasty Generalization".}
\end{center}
\vskip -0.2in
\end{figure}

\cbullet 
Hallucination: Despite using optimized prompts to minimize human error, models occasionally hallucinated. For instance, ChatGPT-4o correctly identified a "False Conversion" fallacy but labeled it with the non-existent term "Fallacy of Conversion."
\vskip 0.05in
\noindent
Ultimately, these misclassifications highlight that LLMs still face significant challenges when performing complex logical reasoning tasks.

\textbf{Instruction-Guided Classification with Relational Graphs} Our analysis of the Instruction-Guided Classification with Relational Graphs method revealed that integrating stepwise instructions with relational graphs provides models with clearer patterns and opportunities for reconsideration. This approach improved performance, with the correct classification frequently appearing as the second-ranked option among misclassified statements (Claude-Sonnet-4: 39\%, ChatGPT-4o: 20\%, Gemini-2.5-Flash: 19\%).

\textbf{Overall Performance and Limitations} The top-performing model achieved only 62\% accuracy, highlighting that LLMs still struggle with complex logical reasoning. A primary cause is their inconsistent adherence to instructions. While prompts instructed models to perform verification checks for every logical fallacy, Claude-Sonnet-4, for example, admitted its initial classifications were often based on intuitive pattern recognition rather than systematic instruction execution. This non-compliance makes subsequent correction unlikely.
Further analysis showed that while decomposing tasks into individual queries could slightly improve accuracy (from 54.70\% to 66.67\% on a 27-example subset), this method proved to be prohibitively resource-intensive and time-consuming, taking two consecutive days for a small sample due to API throttling. 

\textbf{Claude-Sonnet-4 Demonstrated Consistently Superior Performance} 
Across the "Three-tiered Hierarchical Classification," "Stepwise Instructed Classification," and "Instruction-Guided Classification with Relational Graphs" methodologies, Claude-Sonnet-4 consistently outperformed both Gemini-2.5-Flash and ChatGPT-4o, occasionally by substantial margins. In contrast, ChatGPT-4o demonstrated comparatively weaker performance in both the "Stepwise Classification" and "Enhanced Classification" approaches. We initially hypothesized that this performance difference might be attributed to ChatGPT-4o being an earlier-generation model, released approximately one year prior to the other two models. To test this hypothesis, we conducted identical evaluations using ChatGPT-o4-mini, which yielded results nearly identical to those of ChatGPT-4o. These findings suggest that Claude-Sonnet-4 possesses superior logical reasoning capabilities, making it a more suitable candidate for tasks that require advanced logical planning and comprehension.

\textbf{The Logical Fallacy Entrapment} 
Across all methodological approaches, the LLMs demonstrated susceptibility to systematic classification errors, which we term the "Logical Fallacy Entrapment". Our analysis suggests that each model possesses inherent classificatory biases that consistently direct initial fallacy identification toward particular categories. Given the design constraints of our approaches, when initial classifications deviate substantially from the correct labels, subsequent correction becomes virtually impossible. This limitation is particularly evident in the hierarchical design employed in the "Hierarchical Classification" approach, which consequently yielded the poorest performance outcomes across all evaluation approaches.
Furthermore, the models demonstrated unique tendencies in fallacy classification. Claude-Sonnet-4 exhibited a strong preference for Hasty Generalization and the Complex Question Fallacy, which contrasted with ChatGPT-4o's inclination toward Ad Hominem Abusive and the Fallacy of the Undistributed Middle. Gemini-2.5-Flash displayed a different pattern, predominantly choosing Appeal to Emotion and Alleged Certainty.

\textbf{The Paradoxical Effect of Stepwise Instructions and the Corrective Role of Relational Graphs}
The most controversial findings in our study emerge from the "Stepwise Instructed Classification" approach, where Gemini-2.5-Flash and ChatGPT-4o demonstrated diminished performance relative to the baseline, while Claude-Sonnet-4 exhibited improved performance. Upon analyzing these results, we hypothesize that this divergence may be attributed to Claude-Sonnet-4 displaying the least inherent classification bias among all evaluated models and that Claude-Sonnet-4 has better instruction following capabilities. \citep{white2024livebench} Consequently, Claude-Sonnet-4 typically generates more varied initial classifications, thereby increasing the probability of avoiding the systematic classification errors discussed previously.
Notably, all three models demonstrated enhanced performance when provided with both stepwise instructions and relational graphs. These results indicate that by constraining the classification space and compelling LLMs to conduct more thorough analysis, the models achieve superior outcomes even in logically demanding tasks.

\section{Limitations}
\textbf{Prompting Limitation}
We maintained consistent prompt structures across all models within each methodological approach while adhering to established prompting best practices; however, response variability persisted across models. Gemini-2.5-Flash occasionally required multiple prompt iterations to generate valid outputs, frequently producing error messages indicating an inability to access local files or missing classification statements. Moreover, due to inconsistent output formatting from Gemini-2.5-Flash, manual information extraction and post-processing were required.
Similarly, when providing Claude-Sonnet-4 with both stepwise instructions and relational graphs, multi-turn prompting sessions were often required to ensure complete adherence to the specified procedural guidelines.

\textbf{Dataset Limitation}
Our reliance on the FALLACIES dataset resulted in the inheritance of its limitations, including taxonomic overlaps that complicate accurate classification of specific statements. Several logical fallacies exhibit hierarchical relationships where broader categories encompass more specific subcategories. For instance, "Appeal to Emotion" serves as an umbrella category that subsumes more specific fallacies such as "Appeal to Anger" and "Appeal to Fear," given that anger and fear constitute emotional states. Similarly, "Non Sequitur" encompasses various subcategories of logical disconnection.
An additional inherent limitation within the dataset stems from definitional ambiguity, where multiple logical fallacy definitions may legitimately apply to the same statements. This classificatory ambiguity renders certain statements inherently challenging to categorize, presenting difficulties even for human annotators.
While our findings provide initial insights, we acknowledge the limitation that drawing conclusions on LLMs' capabilities in logical fallacy classification from a single data source may not be sufficient. A more comprehensive evaluation should include a broader range of data sources.

\textbf{Data Processing limitation}
An additional limitation concerns the opacity of how LLMs process and utilize the supplementary files we provided, as these internal mechanisms remain inaccessible for examination. Based on our observations, an optimal approach would involve directing LLMs to systematically evaluate each logical fallacy individually to identify all potential matches, followed by a secondary evaluation phase focusing on the narrowed candidate set. However, due to temporal and computational resource constraints, we were unable to implement this iterative approach and instead relied on zero-shot classification methodologies where LLMs managed the entire process autonomously.

\textbf{Relational Graph Construction Bias}
Relational graphs were constructed based on misclassification patterns observed in the baseline results. The underlying hypothesis posits that when LLMs systematically misclassify logical fallacy A as logical fallacy B, these fallacies likely share significant conceptual or structural similarities. However, the baseline statements represent a limited sample that does not encompass all possible classificatory scenarios, resulting in incomplete capture of such relationships within our Prolog relational graphs.

\section{Future Work}
The outcomes of this preliminary investigation suggest several prospective directions for future research.

\textbf{Cross-Model and Cross-Dataset Generalizability Assessment}
The experiments conducted in this initial phase were restricted to closed-source proprietary language models (LMs) and the singular FALLACIES dataset. \citep{hong2024closer} To rigorously assess the generalizability of our methodology, future work will integrate a selection of open-source LMs, such as Vicuna and Llama. Furthermore, we intend to apply our proposed methodology to additional logical fallacy datasets, including the \emph{logical-fallacy} dataset. \citep{jin2022logical} 

\textbf{Fine-tune LMs for Atomic Instruction Following}
Despite the strong performance reported in the literature concerning Large Language Model (LLM) instruction-following capabilities, we faced persistent challenges in strictly enforcing adherence to our atomic instructions. Therefore, a critical future direction involves fine-tuning selected LMs specifically to improve their capability in following atomic instructions. 

\textbf{Enhancement in Knowledge Representation and Graphs}
In this preliminary effort, the relational graph is only used to indicate the common confusion or misclassification between logical fallacies. To facilitate more comprehensive comparison and robust classification, a planned enhancement involves integrating fallacy-specific distinguishing traits into the Prolog-based knowledge graph. This expansion is designed to move beyond merely noting fallacy relationships by integrating step-by-step comparison knowledge for similar fallacies. For example, under the enhanced version of our knowledge graph, distinguishing between \emph{Accent Fallacy} and \emph{Contextomy} hinges on identifying whether the manipulation targets a specific word or the broader contextual framing, enabling Large Language Models (LLMs) to effectively resolve these nuanced cases.

\textbf{Explanation Generation for Classification}
A significant advantage of leveraging LLMs is their capacity for generating coherent explanations. Therefore, another key research trajectory is the development of a system that combines the intermediate outputs from each step in the stepwise instructional reasoning chain to automatically generate an explanation of why a given statement is classified as logical fallacy A rather than logical fallacy B.

\section{Conclusion}
In this research, we sought to answer the question of whether the provision of a structured sequence of stepwise, atomic instructions could induce more deliberate, effortful, and logical reasoning in LLMs. We evaluated four distinct methodological approaches for engaging Claude-Sonnet-4, Gemini-2.5-Flash, and ChatGPT-4o in logical fallacy classification tasks. Our findings demonstrate that the integration of stepwise instructions with Prolog relational graphs effectively constrains the classification space by narrowing the candidate fallacy set and facilitating model reconsideration of initial decisions, thereby yielding improved performance outcomes across all models. However, we also acknowledge that our methodologies are far from perfect and suffer from several limitations, such as prompting, dataset, data processing, and graph construction. A few valuable future directions include expanding the experimental scope by testing the methodology across diverse open-source Language Models and additional logical fallacy datasets; enhancing instruction adherence through fine-tuning LMs specifically for strict atomic instruction following; enriching the knowledge representation by incorporating fallacy-specific traits into the Prolog-based relational graph; and finally, leveraging the LMs' capabilities to generate detailed explanations for classification outcomes based on the stepwise reasoning chain. 
 
\begin{acknowledgements} 
\noindent
This work was partially funded by Underwriters Laboratories Inc. through the Center for Advancing Safety of Machine Intelligence.  This material is based upon work supported by the Air Force Office of Scientific Research under award number FA9550-24-1-0149. 

Tashvi Bansal, Ryan Bai, and Emily M. Chui are equal contributors to this work. We would also like to thank the feedback from the AIEA lab at UC Santa Cruz.
\end{acknowledgements} 

\vspace{-0.25in}

{\parindent -10pt\leftskip 10pt\noindent
\bibliographystyle{cogsysapa}
\bibliography{ref}

\begin{thebibliography}{39}
\expandafter\ifx\csname natexlab\endcsname\relax\def\natexlab#1{#1}\fi
\expandafter\ifx\csname url\endcsname\relax
  \def\url#1{{\path{\sloppy #1}}}\fi
\expandafter\ifx\csname urlprefix\endcsname\relax\def\urlprefix{From }\fi

\bibitem[{Achiam et~al.(2023)}]{achiam2023gpt}
Achiam, J., et~al. (2023).
\newblock Gpt-4 technical report.
\newblock {\em arXiv preprint arXiv:2303.08774\/}.

\bibitem[{{Anthropic Safety \& Research Team}(2025)}]{anthropic2025_claude4_systemcard}
{Anthropic Safety \& Research Team} (2025).
\newblock {\em System card: Claude opus 4 \& claude sonnet 4\/}.
\newblock System card, Anthropic.
\newblock \urlprefix\url{\url{https://www-cdn.anthropic.com/4263b940cabb546aa0e3283f35b686f4f3b2ff47.pdf}}.

\bibitem[{Belle(2025)}]{belle2025relevance}
Belle, V. (2025).
\newblock On the relevance of logic for artificial intelligence, and the promise of neuro-symbolic learning.
\newblock {\em Neurosymbolic Artificial Intelligence\/}, {\em 1\/}, 1--23.

\bibitem[{Bratko(1990)}]{bratko1990prolog}
Bratko, I. (1990).
\newblock Prolog programming for artificial intelligence.

\bibitem[{Chen et~al.(2025)Chen, Zhang, Langrené, \& Zhu}]{chen2025unleashing}
Chen, B., Zhang, Z., Langrené, N., \& Zhu, S. (2025).
\newblock Unleashing the potential of prompt engineering for large language models.
\newblock {\em Patterns\/}, {\em 6\/}, 101260.
\newblock \urlprefix\url{https://www.sciencedirect.com/science/article/pii/S2666389925001084}.

\bibitem[{Chen(2025)}]{chen2025comparative}
Chen, M.~K. (2025).
\newblock A comparative study of neurosymbolic ai approaches to interpretable logical reasoning.
\newblock {\em arXiv preprint arXiv:2508.03366\/}.

\bibitem[{Clark et~al.(1980)Clark, McCabe, \& McCabe}]{clark1980prolog}
Clark, K.~L., McCabe, F., \& McCabe, F. (1980).
\newblock {\em Prolog: a language for implementing expert systems\/}.
\newblock Imperial College of Science and Technology. Department of Computing.

\bibitem[{Colelough \& Regli(2025)}]{colelough2025neuro}
Colelough, B.~C., \& Regli, W. (2025).
\newblock Neuro-symbolic ai in 2024: A systematic review.
\newblock {\em arXiv preprint arXiv:2501.05435\/}.

\bibitem[{Garcez \& Lamb(2023)}]{garcez2023neurosymbolic}
Garcez, A.~d., \& Lamb, L.~C. (2023).
\newblock Neurosymbolic ai: the 3 rd wave.
\newblock {\em Artificial Intelligence Review\/}, {\em 56\/}, 12387--12406.

\bibitem[{Hayes-Roth et~al.(1983)Hayes-Roth, Waterman, \& Lenat}]{hayes1983building}
Hayes-Roth, F., Waterman, D.~A., \& Lenat, D.~B. (1983).
\newblock {\em Building expert systems\/}.
\newblock Addison-Wesley Longman Publishing Co., Inc.

\bibitem[{Hong et~al.(2024)Hong, Zhang, Pang, Yu, \& Zhang}]{hong2024closer}
Hong, R., Zhang, H., Pang, X., Yu, D., \& Zhang, C. (2024).
\newblock A closer look at the self-verification abilities of large language models in logical reasoning.
\newblock {\em Proceedings of the 2024 Conference of the North American Chapter of the Association for Computational Linguistics: Human Language Technologies (Volume 1: Long Papers)\/} (pp. 900--925).

\bibitem[{Hong et~al.(2023)Hong, Zhang, Zhao, Yu, \& Zhang}]{hong2023faithful}
Hong, R., Zhang, H., Zhao, H., Yu, D., \& Zhang, C. (2023).
\newblock Faithful question answering with monte-carlo planning.
\newblock {\em Proceedings of the 61st Annual Meeting of the Association for Computational Linguistics (Volume 1: Long Papers)\/} (pp. 3944--3965).

\bibitem[{Jang et~al.(2023)Jang, Ye, \& Seo}]{jang2023can}
Jang, J., Ye, S., \& Seo, M. (2023).
\newblock Can large language models truly understand prompts? a case study with negated prompts.
\newblock {\em Transfer learning for natural language processing workshop\/} (pp. 52--62). PMLR.

\bibitem[{Jeong et~al.(2025)Jeong, Jang, \& Park}]{jeong2025large}
Jeong, J., Jang, H., \& Park, H. (2025).
\newblock Large language models are better logical fallacy reasoners with counterargument, explanation, and goal-aware prompt formulation.
\newblock {\em Findings of the Association for Computational Linguistics: NAACL 2025\/} (pp. 6918--6937).

\bibitem[{Jiang et~al.(2023)}]{jiang2023followbench}
Jiang, Y., et~al. (2023).
\newblock Followbench: A multi-level fine-grained constraints following benchmark for large language models.
\newblock {\em arXiv preprint arXiv:2310.20410\/}.

\bibitem[{Jin et~al.(2022)Jin, Lalwani, Vaidhya, Shen, Ding, Lyu, Sachan, Mihalcea, \& Schoelkopf}]{jin2022logical}
Jin, Z., Lalwani, A., Vaidhya, T., Shen, X., Ding, Y., Lyu, Z., Sachan, M., Mihalcea, R., \& Schoelkopf, B. (2022).
\newblock Logical fallacy detection.
\newblock {\em Findings of the Association for Computational Linguistics: EMNLP 2022\/} (pp. 7180--7198).

\bibitem[{Kahneman(2011)}]{kahneman2011thinking}
Kahneman, D. (2011).
\newblock {\em Thinking, fast and slow\/}.
\newblock macmillan.

\bibitem[{Kim et~al.(2023)Kim, Park, Jeong, Lee, Han, Lee, \& Kang}]{kim2023better}
Kim, J., Park, S., Jeong, K., Lee, S., Han, S.~H., Lee, J., \& Kang, P. (2023).
\newblock Which is better? exploring prompting strategy for llm-based metrics.
\newblock \urlprefix\url{https://arxiv.org/abs/2311.03754}.

\bibitem[{Kong et~al.(2024)Kong, Zhao, Chen, Li, Qin, Sun, Zhou, Wang, \& Dong}]{kong-etal-2024-better}
Kong, A., Zhao, S., Chen, H., Li, Q., Qin, Y., Sun, R., Zhou, X., Wang, E., \& Dong, X. (2024).
\newblock Better zero-shot reasoning with role-play prompting.
\newblock {\em Proceedings of the 2024 Conference of the North American Chapter of the Association for Computational Linguistics: Human Language Technologies (Volume 1: Long Papers)\/} (pp. 4099--4113). Mexico City, Mexico: Association for Computational Linguistics.
\newblock \urlprefix\url{https://aclanthology.org/2024.naacl-long.228/}.

\bibitem[{Kroening \& Strichman(2016)}]{kroening2016decision}
Kroening, D., \& Strichman, O. (2016).
\newblock {\em Decision procedures\/}, volume~1.
\newblock Springer.

\bibitem[{Kumar et~al.(2024)}]{kumar2024training}
Kumar, A., et~al. (2024).
\newblock Training language models to self-correct via reinforcement learning.
\newblock {\em arXiv preprint arXiv:2409.12917\/}.

\bibitem[{Lei \& Huang(2024)}]{lei2024boosting}
Lei, Y., \& Huang, R. (2024).
\newblock Boosting logical fallacy reasoning in llms via logical structure tree.
\newblock {\em Proceedings of the 2024 Conference on Empirical Methods in Natural Language Processing\/} (pp. 13157--13173).

\bibitem[{Li et~al.(2023)Li, Jansen, Huang, Lee, Ganti, \& Kuzmin}]{li2023maqa}
Li, J.~Y., Jansen, A., Huang, Q., Lee, J., Ganti, R., \& Kuzmin, D. (2023).
\newblock Maqa: A multimodal qa benchmark for negation.
\newblock {\em arXiv preprint arXiv:2301.03238\/}.

\bibitem[{Li et~al.(2025)}]{li2025system}
Li, Z.-Z., et~al. (2025).
\newblock From system 1 to system 2: A survey of reasoning large language models.
\newblock {\em arXiv preprint arXiv:2502.17419\/}.

\bibitem[{Liu et~al.(2023)Liu, Ning, Teng, Liu, Zhou, \& Zhang}]{liu2023evaluating}
Liu, H., Ning, R., Teng, Z., Liu, J., Zhou, Q., \& Zhang, Y. (2023).
\newblock Evaluating the logical reasoning ability of chatgpt and gpt-4.
\newblock {\em arXiv preprint arXiv:2304.03439\/}.

\bibitem[{Lou et~al.(2024)Lou, Zhang, \& Yin}]{lou2024large}
Lou, R., Zhang, K., \& Yin, W. (2024).
\newblock Large language model instruction following: A survey of progresses and challenges.
\newblock {\em Computational Linguistics\/}, {\em 50\/}, 1053--1095.

\bibitem[{Luettgau et~al.(2025)Luettgau, Kirk, Hackenburg, Bergs, Davidson, Ogden, Siddarth, Huang, \& Summerfield}]{luettgau2025conversational}
Luettgau, L., Kirk, H.~R., Hackenburg, K., Bergs, J., Davidson, H., Ogden, H., Siddarth, D., Huang, S., \& Summerfield, C. (2025).
\newblock Conversational ai increases political knowledge as effectively as self-directed internet search.
\newblock {\em arXiv preprint arXiv:2509.05219\/}.

\bibitem[{Lund \& Villadsen(2022)}]{lund2022verified}
Lund, S.~T., \& Villadsen, J. (2022).
\newblock On verified automated reasoning in propositional logic.
\newblock {\em Asian Conference on Intelligent Information and Database Systems\/} (pp. 390--402).

\bibitem[{Mishra et~al.(2021)Mishra, Khashabi, Baral, Choi, \& Hajishirzi}]{mishra2021reframing}
Mishra, S., Khashabi, D., Baral, C., Choi, Y., \& Hajishirzi, H. (2021).
\newblock Reframing instructional prompts to gptk's language.
\newblock {\em arXiv preprint arXiv:2109.07830\/}.

\bibitem[{Pan et~al.(2023)Pan, Albalak, Wang, \& Wang}]{pan2023logic}
Pan, L., Albalak, A., Wang, X., \& Wang, W. (2023).
\newblock Logic-lm: Empowering large language models with symbolic solvers for faithful logical reasoning.
\newblock {\em Findings of the Association for Computational Linguistics: EMNLP 2023\/} (pp. 3806--3824).

\bibitem[{Qin et~al.(2024)}]{qin2024infobench}
Qin, Y., et~al. (2024).
\newblock Infobench: Evaluating instruction following ability in large language models.
\newblock {\em Findings of the Association for Computational Linguistics ACL 2024\/} (pp. 13025--13048).

\bibitem[{Saccon et~al.(2024)Saccon, Tikna, De~Martini, Lamon, Palopoli, \& Roveri}]{saccon2024prolog}
Saccon, E., Tikna, A., De~Martini, D., Lamon, E., Palopoli, L., \& Roveri, M. (2024).
\newblock When prolog meets generative models: a new approach for managing knowledge and planning in robotic applications.
\newblock {\em 2024 IEEE International Conference on Robotics and Automation (ICRA)\/} (pp. 17065--17071). IEEE.

\bibitem[{Saunders et~al.(2022)Saunders, Yeh, Wu, Bills, Ouyang, Ward, \& Leike}]{saunders2022self}
Saunders, W., Yeh, C., Wu, J., Bills, S., Ouyang, L., Ward, J., \& Leike, J. (2022).
\newblock Self-critiquing models for assisting human evaluators.
\newblock {\em arXiv preprint arXiv:2206.05802\/}.

\bibitem[{Tan et~al.(2024)Tan, Deng, Qiu, Xu, Qu, Chu, Xu, \& Qi}]{tan2024thought}
Tan, X., Deng, Y., Qiu, X., Xu, W., Qu, C., Chu, W., Xu, Y., \& Qi, Y. (2024).
\newblock Thought-like-pro: Enhancing reasoning of large language models through self-driven prolog-based chain-of-thought.
\newblock {\em arXiv preprint arXiv:2407.14562\/}.

\bibitem[{Team et~al.(2023)}]{team2023gemini}
Team, G., et~al. (2023).
\newblock Gemini: a family of highly capable multimodal models.
\newblock {\em arXiv preprint arXiv:2312.11805\/}.

\bibitem[{Vakharia et~al.(2024)Vakharia, Kufeldt, Meyers, Lane, \& Gilpin}]{vakharia2024proslm}
Vakharia, P., Kufeldt, A., Meyers, M., Lane, I., \& Gilpin, L.~H. (2024).
\newblock Proslm: A prolog synergized language model for explainable domain specific knowledge based question answering.
\newblock {\em International Conference on Neural-Symbolic Learning and Reasoning\/} (pp. 291--304).

\bibitem[{White et~al.(2024)}]{white2024livebench}
White, C., et~al. (2024).
\newblock Livebench: A challenging, contamination-limited llm benchmark.
\newblock {\em arXiv preprint arXiv:2406.19314\/}.

\bibitem[{White et~al.(2023)White, Fu, Hays, Sandborn, Olea, Gilbert, Elnashar, Spencer-Smith, \& Schmidt}]{white2023prompt}
White, J., Fu, Q., Hays, S., Sandborn, M., Olea, C., Gilbert, H., Elnashar, A., Spencer-Smith, J., \& Schmidt, D.~C. (2023).
\newblock A prompt pattern catalog to enhance prompt engineering with chatgpt.
\newblock {\em Proceedings of the 30th Conference on Pattern Languages of Programs\/} (pp. 1--31).

\bibitem[{Zamfirescu-Pereira et~al.(2023)Zamfirescu-Pereira, Wong, Hartmann, \& Yang}]{zamfirescu2023johnny}
Zamfirescu-Pereira, J.~D., Wong, R.~Y., Hartmann, B., \& Yang, Q. (2023).
\newblock Why johnny can’t prompt: how non-ai experts try (and fail) to design llm prompts.
\newblock {\em Proceedings of the 2023 CHI conference on human factors in computing systems\/} (pp. 1--21).

\end{thebibliography}

}

\newpage
\appendix
\section{Prompts}
\subsection{Prompt for Baseline Establishment}
If you have to choose one classification among all the level 3 labels, which one will you pick for this statement? 
Below are all the level 3 labels and their descriptions.

\subsection{Prompts Employed in Three-Tiered Hierarchical Classification}
\subsubsection{First Tier Classification}
I will ask you to classify the specific type of logical fallacies in the example. You will get a different reward for getting the correct answer on different levels. For example, if you can get the correct classification on level 1, you get reward 1; otherwise, you get 0. If you can get the correct classification on level 1, you move on to the next level, and if you can get the correct classification on level 2, you get reward 2, if you can get the correct classification on level 3, you get reward 3, and so on. The first level of classification challenge is to classify whether the logical fallacy in the statement is formal or informal. The definition of a formal fallacy means that there is an error in the argument's form. In contrast, an informal fallacy means the arguments are logically unsound for a lack of well-grounded premises. Now classify the below statement and output the results in this format: "level\_1\_results:\#{}\#" in all lower cases with no "fallacy". Only output the level 2 classification, no explanations.

\subsubsection{Second Tier Classification}
Now you are moved to level 2 of classification, where you need to further classify the fallacy. Under formal fallacy, there are Proposition(Errors in dealing with the logical relations holding between propositions.), Quantification Fallacy (Errors in dealing with the quantifiers), Syllogism Fallacy (Errors in the syllogisms of deductive reasoning.), Probability Fallacy (Errors in dealing with probability.). Under informal fallacy, there are Ambiguity Fallacy (Errors due to linguistic ambiguity or vagueness of terms), Inconsistency Fallacy (Self-contradiction and inconsistency occur). Irrelevance Fallacy (The premises are irrelevant to the conclusion.), Insufficiency Fallacy (The premises are insufficient or weak to support the conclusion.), and Inappropriate Presumption Fallacy (An inappropriate presumption is explicitly or implicitly introduced.), which one do you think it would fall under? You can also go back and change your previous classification. Output the results in this format: "level\_2\_results:\#{}\#" in all lower cases with no "fallacy". Only output the level 2 classification, no explanations. Only output a classification from the level 2 fallacy list provided in the format requested.

\subsubsection{Sample Third-Tier Classification Prompt within the Syllogism Category}
Moving on to level three, under Syllogism, there are Fallacy of the Undistributed Middle (A formal fallacy in a categorical syllogism where the middle term, or the term that does not appear in the conclusion, is not distributed to the other two terms.), Exclusive Premises (A standard form categorical syllogism that has two negative premises either in the form of “no X are Y” or “some X are not Y”), Fallacy of Four Terms (This fallacy occurs in a categorical syllogism when the syllogism has four terms rather than the requisite three.), Illicit Substitution of Identicals (A fallacy due to confusing the knowing of a thing (extension) with the knowing of it under all its various names or descriptions (intension).), Illicit Minor (Any form of a categorical syllogism in which the minor term is distributed in the conclusion, but not in the minor premise.), Illicit Major (Any form of a categorical syllogism in which the major term is distributed in the conclusion, but not in the major premise.), Negative Conclusion from Affirmative Premises (The conclusion of a standard form categorical syllogism is negative, but both of the premises are positive.), Affirmative Conclusion from a Negative Premise (The conclusion of a standard form categorical syllogism is affirmative, but at least one of the premises is negative). Which one do you think it should fall under? You can also go back and change your previous classification. Output the results in this format: "level\_3\_results:\#{}\#" in all lower cases with no "fallacy". Only output the level 3 classification, no explanations. Only output a classification from the level 3 fallacy list provided in the format requested.

\subsection{Prompt for Stepwise Instructed Classification}
You are an expert at classifying logical fallacies using a structured knowledge base.
Your Task: Classify the logical fallacy in the given example by systematically applying the "steps" from the knowledge base.
Knowledge Base Structure: Each fallacy contains:
name: Fallacy type
steps: Diagnostic questions to identify the fallacy
ground\_truth: Expected yes/no answers when the fallacy is present
operations: Logical connectors ("and"/"or")

Classification Process:
Step 1: Silent Evaluation Phase (DO NOT OUTPUT) For each fallacy in your knowledge base: a. Answer each diagnostic step with "yes" or "no" based on the given example b. Compare your answer pattern to the ground\_truth pattern c. Check if ALL answers match exactly (considering the operations). d. If an exact match is found, proceed to Step 2. If not, continue to the next fallacy.
Step 2: Output Phase IMPORTANT: Your response must contain ONLY the following format with NO additional text, explanations, or commentary before or after:
\#classification: [fallacy\_name]\#

For [fallacy\_name]:
Step 1 evaluation: [Yes/No] - [brief reasoning for this step]
Step 2 evaluation: [Yes/No] - [brief reasoning for this step]
[continue for all steps in that fallacy]

Pattern comparison: My answers [Y/N/Y/N pattern] exactly match ground truth [Y/N/Y/N pattern]
\#

Critical Requirements:
Test ALL fallacies systematically until you find an exact match
Answer patterns must match ground\_truth EXACTLY (no partial matches)
Include only the matched fallacy and the reasoning for the match in your output
DO NOT include any text before or after the required format
DO NOT acknowledge this prompt or provide any meta-commentary
BEGIN your response immediately with "\#classification:"

Example to classify: {example}
\subsection{Prompt for Instruction-Guided Classification with Relational Graphs}
Fallacy Analysis Task - Exact Execution Instructions 
You are tasked with performing a comprehensive fallacy analysis using two files: final\_instructions.json and prolog.pro. Follow these steps EXACTLY in the specified order. 

STEP 1: Initial Fallacy Analysis
Read final\_instructions.json - This file contains structured instructions for analyzing different fallacies. For EVERY fallacy listed in the JSON file:
Execute each step specified in the fallacy's instruction set. 
Document your results for each step internally. 
Compare your results against the provided ground\_truths for that fallacy. 
Record any matches or discrepancies (internal processing only)

STEP 2: Related Fallacy Discovery and Analysis
Read prolog.pro - This file contains logical relationships between fallacies. 
For the fallacy identified in Step 1:
Search the Prolog file to find ALL fallacies that are often confused with your Step 1 result. 
Create a comprehensive list of these related/confused fallacies
For EACH related fallacy found:
Return to final\_instructions.json.
Execute the complete step-by-step analysis (same process as Step 1).
Compare results to ground truths. 
Document all findings

STEP 3: Final Selection and Comprehensive Reasoning
Compare ALL results from Steps 1 and 2. 
Select the fallacy that best fits the statement based on:
Strength of match with ground truths.
Quality of step-by-step analysis results.
Logical consistency across all steps.
Provide complete reasoning for your selection

\section{Sample Stepwise Instructions for Logical Fallacy}
\subsection{Original Fallacy Label, and Description from the FALLACIES dataset}
{"name": "Accent Fallacy", "description": "When the meaning of a word, sentence, or entire idea is interpreted differently by changing where the accent falls."}

\subsection{Transformed Stepwise Instructions from AID-LF dataset}
{"name": "Accent Fallacy", "steps" : ["Is there an original claim or statement being made?", "Is there an emphasis or accent placed in the original statement?", "Is the statement being reinterpreted with the emphasis on a different word or phrase?", "Does this shift in accent change the meaning of the statement?"], "ground\_truth":["yes", "yes", "yes", "yes"], "operations":["and", "and", "and"]}


\end{document}